\documentclass[lettersize,journal]{IEEEtran}

\usepackage{cite,pifont}
\usepackage{amsmath,amssymb,amsfonts}
\usepackage{subfiles,amsthm}
\usepackage{textcomp}
\DeclareMathAlphabet{\pazocal}{OMS}{zplm}{m}{n}
\newcommand{\Lb}{\pazocal{L}}
\newcommand{\Lt}{\pazocal{T}}
\newcommand{\Ln}{\pazocal{N}}
\newcommand{\Lv}{\pazocal{V}}
\newcommand{\Lcp}{\pazocal{CP}}
\usepackage{graphicx}

\usepackage{algpseudocode}
\usepackage[caption=false,font=small,labelfont=sf]{subfig}
\usepackage{tikz}

\usepackage{multirow}
\usepackage{xcolor}
\usepackage[linesnumbered,ruled,vlined]{algorithm2e}

\SetCommentSty{mycommfont}
\usepackage[super]{nth}
\SetKwInput{KwInput}{Input}                
\SetKwInput{KwOutput}{Output}   

\newlength{\dhatheight}

    \newcommand{\mycomment}[1]{}
    
    \newtheorem{lemma}{Lemma}
\newtheorem{example}{Example}
\newcommand{\myparagraph}[1]{\vspace{3.0pt}\noindent{\textbf{#1.}}~}
\newtheorem{definition}{Definition}

\def\BibTeX{{\rm B\kern-.05em{\sc i\kern-.025em b}\kern-.08em
    T\kern-.1667em\lower.7ex\hbox{E}\kern-.125emX}}
    \newcommand\method[1]{{\mbox{\sf\small{#1}}}}
    \newcommand{\newprob}{\method{PERP}\xspace}

\begin{document}

\title{Proactive Route Planning for Electric Vehicles}

\author{\IEEEauthorblockN{Saeed Nasehi, Farhana Choudhury, Egemen Tanin}%
\thanks{Manuscript received - -, 2024; revised - -, 2024.}
\thanks{Saeed Nasehi, Farhana Choudhury, and Egemen Tanin are with the School of Computing and Information Systems, The University of Melbourne, Melbourne, VIC 3052, Australia (e-mails:
\{s.nasehibasharzad@student.,farhana.choudhury@,etanin@\}unimelb.edu.au).}
}

\markboth{}%
{Nasehi \MakeLowercase{\textit{et al.}}: Proactive Route Planning for Electric Vehicles}%
\maketitle
\begin{abstract}
Due to the limited driving range, inadequate charging facilities, and time-consuming recharging, the process of finding an optimal charging route for electric vehicles (EVs) differs from that of other vehicle types. The time and location of EV charging during a trip impact not only the individual EV's travel time but also the travel time of other EVs, due to the queuing that may arise at the charging station(s). This issue is at large seen as a significant constraint for uplifting EV sales in many countries. In this study, we present a novel Electric Vehicle Route Planning problem, which involves finding the fastest route with recharging for an EV routing request. We model the problem as a new graph problem and present that the problem is NP-hard. We propose a novel two-phase algorithm to traverse the graph to find the best possible charging route for each EV. We also introduce the notion of `influence factor' to propose heuristics to find the best possible route for an EV with the minimum travel time that avoids using charging stations and time to recharge at those stations which can lead to better travel time for other EVs. The results show that our method can decrease total travel time of the EVs by 50\% in comparison with the state-of-the-art on a real dataset, where the benefit of our approach is more significant as the number of EVs on the road increases.
\end{abstract}

\begin{IEEEkeywords}
Electric Vehicle, Route Planning, Partial Recharging 
\end{IEEEkeywords}

\vspace{-11pt}
\section{Introduction}

There has been a surge in the uptake of electric vehicles (EVs) in recent years, and number is expected to grow at an accelerated rate in near future. Moreover, the recent  advancements have led to an increase in EV's battery capacity, making EVs a potential alternative for traditional vehicles~\cite{basharzad2022electric}. Typically, private vehicles are one of the preferred modes of transportation for long trips. According to~\cite{bts_report}, in the US over the year 2021, almost 800 million trips with distance of over 300 miles have been undertaken by private vehicles.
However, long distance trips require one or multiple recharging along the way. As any queuing in the Charging Stations (CSs) can rapidly accumulate to a substantial wait time~\cite{basharzad2022electric} due to prolonged recharging time of EVs\footnote{Even with a supercharger of 150kW, it takes 30 minutes to charge a Tesla-3 model to obtain the range of approximately 550km (80\% charged), source: https://pod-point.com/guides/vehicles/tesla/2021/model-3}, taking long trips by EVs will be impractical without careful planning when the number of EVs sharply increase on the road. This calls for solutions that not only find optimal charging routes with minimal travel time, but also reduce congestion (that is, queuing) at the CSs. 

While there are existing studies on EV route planning with recharging, they suffer from the following two drawbacks: 

\noindent \textbf{Drawback 1 - Inability to find charging route for an EV with the minimum travel time that considers multiple and potentially partial recharging:} While there are recent studies to find charging routes for EVs to minimize the travel time, the studies suffer from at least one or both of the issues from the following, (a) they find only the next best CS to charge~\cite{qian2019deep,lee2020deep,zhang2020effective}; (b) they only consider full recharge at each CS~\cite{qian2019deep,zhang2020effective,gareau2019efficient,de2015intention,qin2011charging,cao2019electric}. 

For long trips, vehicles may require multiple recharges due to their limited driving range. Even for distances that can be covered by a single recharge, the optimal route with the minimum total travel time might involve multiple partial recharges due to different charging rates of the CSs, which is not addressed by existing solutions that focus on iteratively finding the next best CS or solely on full recharges. The only approach that considers partial recharging~\cite{zhang2018shortest} at multiple CSs
demonstrates the potential for routes with up to 45\% less total travel time compared to those ignoring partial recharging. However, this approach neglects waiting time at CSs, a significant factor in realistic EV route planning according to other studies. Therefore, a comprehensive solution is required to find charging routes with the minimum travel time, incorporating both wait time and multiple partial recharging options, to achieve optimal travel time for each EV.

\noindent \textbf{Drawback 2 - No attempt at reducing congestion at charging stations:} All the above mentioned existing studies solely focus on finding an optimal charging route for each EV individually, without any consideration for other EVs on the road. While some research delves into finding a global optima for a group of EVs to minimize the total travel of the group~\cite{qin2021review,yuan2019p}, these solution may generate routes with significantly longer travel times for some EVs to achieve a shorter average travel time overall. Such routes are unsuitable for independent drivers as they have no shared objectives and only prioritize reaching their destinations in the shortest time possible. Moreover, EVs arrive as a stream on the road, making it impractical to know all the EVs' origins and destinations beforehand to form such a realistic group. However, such global optimization approaches better utilize the charging resources by reducing congestion at CSs (that is, less waiting time), leading to decreased waiting times for EVs overall.

\begin{figure}[t]
\centering
 \includegraphics[width=0.49\linewidth]{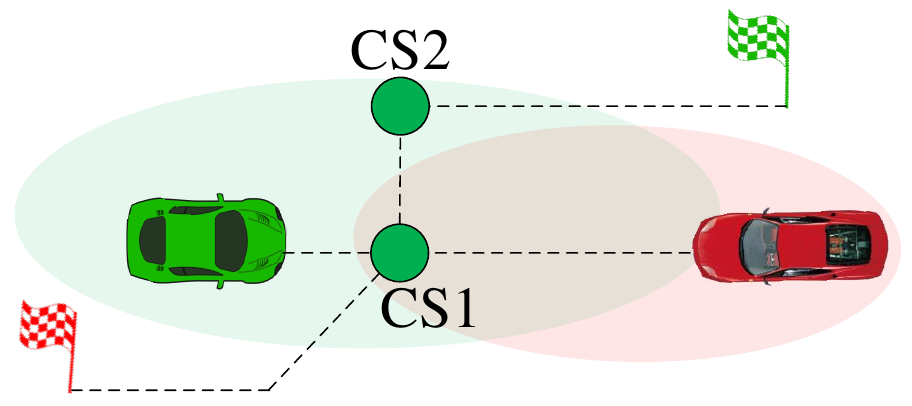}
 \vspace{-10pt}
  \caption{The origin, destination, and driving range of two EVs, along with the locations of CSs are shown. If the green EV charges at CS1, then the travel time of the red EV gets extended, as the red EV has to wait until CS1 becomes available. If the green EV charges at CS2 instead of CS1, that results into a shorter travel time for the red EV. Both paths have the same travel time for the green EV.}
  \vspace{-22pt}
\label{fig:PERPExample}
\end{figure}

\emph{In this paper, our objective is to find a solution that addresses both drawbacks.} To address drawback 1, we aim to find optimal routes for each EV with one or multiple partial recharging at CSs, such that the routes have the minimum total travel time (including waiting time at CSs) for that EV. To address drawback 2, we propose a novel proactive aspect of the problem, where the key idea is to select the route from the optimal charging routes for each EV such that, the selected route avoids the CSs that can be better utilized by upcoming EVs as much as possible (the second route for the green EV in Figure~\ref{fig:PERPExample}, charging at CS2 instead of CS1). This concept is grounded in the observation that there can be more than one optimal charging route for an EV with the same minimum travel time, with different amounts to charge at different CSs, even along the same highway. \textbf{In essence, our novel problem is to ensure the optimality of each EV's charging route individually, while refrain from using the CSs which can be better utilized by other EVs, hence, improving the travel time of all EVs. We denote this problem as Proactive EV Route Planning (\newprob) problem.} 

\vspace{-1pt}
Devising an efficient and scalable solution of the problem is challenging for the following reasons: (i) \textbf{Hardness of the problem:} Even finding a route with the minimum travel time for each EV with one or multiple recharging, which also requires finding the optimal amount of time to charge (i.e., partial recharge) at CS(s), is NP-hard (shown in Section~\ref{sec:model}). (ii) \textbf{Streaming nature:} As EVs arrive as a stream, it is not straightforward to find a route that will minimize the wait time for others, without knowing the EVs that will arrive in future. (iii) \textbf{Efficiency and scalability issues:} Developing a practical route-planning solution is complicated due to the NP-hard nature of the problem and the requirement to minimize wait times for other EVs. Additionally, the solution requires to be scalable for a large number of CSs, as there can be many options to charge along a long trip and we aim to identify the route that minimizes the wait time for future upcomming EVs. Many of the existing approaches for EV routing even without partial recharging and future requests consideration suffer from efficiency and scalability issue (details in Section~\ref{sec:rel}).

In summary, our contributions include: (i) We propose and formalize the novel problem of Proactive EV Route Planning for a stream of EVs, aimed at minimizing the travel time for the current EV in a stream while proactively avoiding CSs that could be better utilized by upcoming EVs as much as possible. (ii) We model the problem as a graph problem with time-dependent self loops to capture both the spatial and temporal aspects of the problem within the graph itself (Section~\ref{sec:model}). This graph model is easily generalizable to many different versions of EV charging problem studied in literature, including different energy consumption rates, charging rates, etc (Section~\ref{Discussion}). (iii) We propose a two-phase efficient algorithm to find the optimal charging route of EVs with the minimum travel time, that can include one or multiple partial recharges to \textbf{address Drawback 1} (Section~\ref{ProactiveAlgorithm}). (iv) We propose `influence factor' and several different heuristics to find a route for the current EV that can reduce travel time for future EVs to \textbf{address Drawback 2} (Section~\ref{ProactiveAlgorithm}). (v) Our experimental results on a real dataset show that our approach can reduce the total travel time by 50\% compared to the state-of-the-art baseline, and the benefit is higher for larger number of EVs. The trade-offs of efficiency and reduction in travel time for future EVs by different heuristics are also evaluated in the experiments and thoroughly discussed (Section~\ref{sec:exp}).

\section{Related work}\label{sec:rel}

EV route planning problem has been studied in recent years, for both individual vehicles and fleets. For individual vehicles, the focus is on finding optimal routes independently, without shared objectives among EVs. For a fleet of vehicles, the objective is to find a route for each vehicle on the fleet based on a shared objective that is optimal for the overall fleet.

\begin{table*}[h!]
\centering
\caption{Summary of closely related studies}
\vspace{-3pt}
\begin{tabular}{lll l l} 
 \hline
 Reference & Multiple recharging & Partial recharging & Optimization objective & Wait time consideration\\
 \hline
 \cite{qian2019deep} & \text{\sffamily X} &  \text{\sffamily X} & Individual & Approximate \\ 
 \hline
 \cite{lee2020deep} & \text{\sffamily X} & Only enough for next CS & Individual & Anticipation - Reservation \\ \hline
 \cite{zhang2020effective} & \text{\sffamily X} & \text{\sffamily X} & Individual & Queue - FCFS \\
 \hline
\cite{schoenberg2022reducing} & \checkmark  & Only enough for next CS & Individual &  Queue - FCFS \\
\hline
Our approach & \checkmark & \checkmark & Proactive & Queue - Reservation\\
 \hline
\end{tabular}
\vspace{-15pt}
\label{table:summary}
\end{table*}

\vspace{-8pt}
\subsection{Individual EV route planning}
The primary distinction between EV and conventional route planning lies in the limitation of an EV’s battery, a constraint that makes EV routing to be treated as two different optimization problems: Energy consumption optimization~\cite{eisner2011optimal,sachenbacher2011efficient,baum2013energy,zhou2019energy}, and travel time optimization~\cite{goodrich2014two,storandt2012quick,zhang2018shortest,schoenberg2022reducing}. We focus on the EV travel time optimization studies as they are closely related to our objective. 
Table~\ref{table:summary} summarizes the most relevant studies and highlights the distinctions from our approach.

The travel time optimization for EVs requires considering refueling at CSs. While some studies neglect waiting time (i.e., no queuing) at CSs~\cite{storandt2012quick,morlock2019time, zhang2018shortest,liu2023optimal,zhang2021bis4ev,goodrich2014two}, it has been shown that waiting time significantly impacts EV route planning~\cite{ben2022route,qian2019deep,schoenberg2022reducing,lee2020deep,zhang2020effective,de2015intention, qin2011charging,bedogni2016route,cao2019electric,hou2019bidding,gareau2019efficient}. The studies on the travel time optimization for EVs with recharging that consider the waiting time can be categorized based on the number of recharges allowed along the route, i.e., single vs. multiple recharges. 

\subsubsection{\textbf{Route planning with single recharging}} 
The studies in this category present solutions to find only the next optimal CS from the current location to recharge~\cite{qian2019deep, lee2020deep,zhang2020effective,ben2022route}, assuming that the EVs at the chosen CS will either fully recharge~\cite{qian2019deep,zhang2020effective,ben2022route}, or charge enough to reach to their destination~\cite{lee2020deep}. 
These approaches are suitable for urban settings where the distance to a CS from any location is not large. For long trips, EVs may require multiple recharges, and the distance between CSs can be substantial on highways. Even for the distances within the range of a single recharge, the optimal route with the minimum total travel time may consist of multiple partial recharges as shown in~\cite{zhang2018shortest}, which cannot be addressed by iteratively finding only the next best CS to charge. Moreover, scalability issues arise in these approaches, especially when confronted with a large number of CSs. The solution in~\cite{ben2022route} uses NSGA-II, a genetic algorithm based solution that exhibits significant computation time, particularly for a large number of CSs. The authors in~\cite{qian2019deep, lee2020deep} apply deep reinforcement learning to solve the problem. Due to the large state and action space, these solutions face scalability and efficiency challenges~\cite{zhang2022rlcharge} even with a small number of CSs. For
instance, both \cite{qian2019deep, lee2020deep} performed their experiments with only 3 CSs.

\subsubsection{\textbf{Route planning with multiple recharging}} Some studies that consider multiple recharging assume that the EV fully recharges at each of those CSs~\cite{gareau2019efficient,de2015intention,qin2011charging,cao2019electric}. Schoenberg et al.~\cite{schoenberg2022reducing} proposed a hybrid approach for recharging -- full charge when the next charging stop has a lower charging rate than the current CS, otherwise, enough charge to reach to the next CS. The only approach that considers partial recharging~\cite{zhang2018shortest} at multiple CSs shows that they can find routes with up to 45\% less total travel time compared to the ones that do not consider partial charging. However, no waiting time is considered in~\cite{zhang2018shortest}. 

We find routes for EVs with one or more partial recharging with the minimum total travel time including waiting time at CSs, which has not been addressed in literature. Moreover, our approach differs significantly from related works as we not only find the optimal path for individual EVs, but also consider the `impact' of an EV's route on potential future EV routing requests. Our goal is to identify all paths guaranteed to be optimal for each individual EV (as if an EV has no shared objective with other EVs), and proactively choose the path that maintains the availability of CSs and/or timeslots that can be optimally utilized by others.

\vspace{-9pt}
\subsection{Route planning for a fleet of EVs}
The navigation and charging problem for a fleet of EVs, particularly e-taxis, have been explored in literature~\cite{yuan2019p,wang2021data,dong2017rec,tu2020real}. These studies focus on optimizing real-time navigation for efficient passenger pickups and drop-offs, incorporating vehicle charging to maintain operational efficiency and service availability. Reducing the queuing time~\cite{dong2017rec}, minimizing the traveling time to CSs~\cite{yuan2019p}, and fairness-aware scheduling~\cite{wang2021data} for e-taxis are some of the studied objectives. In contrast, our aim is to find a route for individual EVs such that each EV gets the optimal path in terms of travel time, while minimizing the total travel time for all EVs in the system. 

\vspace{-9pt}
\subsection{Consideration for reservations}
It is worth noting that reservation systems can be incorporated into EV route planning. In the context of route planning with recharging, some studies~\cite{schoenberg2022reducing,zhang2020effective,gareau2019efficient} assume a first-come-first-serve (FCFS) approach at CSs, while others~\cite{lee2020deep,qin2011charging,cao2019electric,hou2019bidding,de2015intention} employ reservation system to manage and calculate waiting times.  FCFS-based solutions may lead to extended travel times due to unreliable wait time estimates and the need for frequent rerouting. In contrast, reservation-based routing systems have demonstrated a substantial reduction in total travel time, offering accurate wait time estimates and eliminating the need for rerouting. Hence, we follow the reservation system setting in this paper, ensuring precise wait time estimates without requiring trip declarations in advance.

\vspace{-9pt}
\subsection{Other route planning}\label{subsec:other_related}
Some route planning studies focus on alleviating congestion by devising routes with multiple alternatives~\cite{amores2024flexible} or employing top-k shortest path algorithms~\cite{chondrogiannis2020finding,yu2024distributed}, rather than strictly adhering to the shortest path. There are also some studies on energy-efficient EV route planning that are modeled as a Constrained Shortest Path (CSP) problem~\cite{artmeier2010optimal,zhou2019energy}. While these approaches minimize overall travel time or energy consumption (some are in dynamic traffic conditions), they lack applicability to \newprob, as the inclusion of CSs in \newprob introduces unique challenges, such as determining when, where, and how much to recharge, along with calculating associated wait times which surpass the capabilities of such solutions.

\section{Preliminaries}

\mycomment{
\begin{table}[t]
\centering
\caption{Notations and Explanations.}
\begin{tabular}{c | c} 
 \hline
 Notation & Explanations \\ [0.5ex] 
 \hline
 $r$ & A routing request \\
 $R$ & Set of routing requests \\
 $p$ & Charging path \\
 $OCP(r)$ & Set of optimal charging path for request $r$ \\
 $C$ & Set of charging stations in the graph \\
 $t_r$ & Time of request \\
 $\eta_c$ & Charging rate of charging station $c$\\
 $vcr_r$ & vehicle charging rate of request $r$ \\
 $tt(p,r)$ & Total travel time using charging path $p$ for request $r$\\
 $B_r$ & Maximum battery capacity of request $r$\\
 $t^-_c$ & Arrival time to station $c$ \\
 $t^+_c$ & Leave time of station $c$ \\
 $SoC^-_c$ & State of Charge when reach to station $c$ \\
 $SoC^+_c$ & State of Charge when leave station $c$\\
 \hline
\end{tabular}
\label{table:abbr}
\end{table}
}

Let $C$ be a set of charging stations, each $c \in C$ is associated with its spatial location and charging rate $\eta_c$. Now consider a stream of EV routing requests $(r_1, r_2, \dots)$ in order of their request time. Each request $r$ includes the request time (travel start time) $t_r$, origin and destination spatial locations presented as $o_r$ and $d_r$ respectively, vehicle charging rate $\Gamma_r$, maximum battery capacity $B_r$, and initial state of charge $SoC_r$ (as a percentage). We assume a reservation system is implemented in each CS, enabling EVs to schedule their charging. In a CS $c$, one or multiple consecutive timeslots of a fixed length (e.g., 5 minutes) are needed to be reserved by an EV request $r$ in order to get charged in $c$ at those reserved timeslots.

{\definition{Optimal Charging Path (OCP) with Multiple Partial Recharging.}} Given a routing request $r$, an EV route planning with multiple partial recharging is to find a sequence of one or more CSs from $C$ and the charging time at each of those station(s) (i.e., the timeslots to reserve at each CS in that sequence), so that, the total time from $o_r$ to $d_r$ is minimum, without running out of charge (i.e., always $SoC_r>0$ until destination). We denote such a sequence with the minimum total time as `optimal charging path, OCP(r)'. Note that, there can be more than one OCP for a request with the same travel time. The total travel time $tt(p,r)$ for a charging path $p \in OCP(r)$, from $o_r$ to $d_r$, is the summation of the driving time $dt$, charging time $ct$, and waiting time $wt$ at the charging station(s), calculated as,

\vspace{-15pt}
\begin{equation}
\begin{split}
\vspace{-5pt}
tt(p,r)= dt(o_r,c_1) + \Bigg( \sum^{|p|-1}_{i=1} dt(c_i,c_{i+1})  + ct(c_i) \\  +wt(c_i) \Bigg) + dt(c_{|p|-1},d_r)
\end{split}
\end{equation}
\vspace{-10pt}

\myparagraph{Effective charging rate} The charging rate of a CS refers to the station's ability to supply electric power to the EV, commonly measured in kilowatt hour (kWh). Nonetheless, this value is not sufficient for accurately calculating the charging time of an EV at a CS, as the vehicle charging rate specific to each vehicle serves as an additional influential factor. 

The effective charging rate $CR(c,r)$ for an EV in a CS is determined as the minimum value between the CS's charging rate $\eta_{c}$ and the EV's receiving charging rate $\Gamma_r$~\cite{morlock2019time, Zapmap-charging-connector}, calculated as $CR(c,r) = min (\eta_{c}, \Gamma_r)$. Take an EV with $\Gamma_r= 50$ kWh as an example. If it encounters three CSs with rates ($\eta_c$) of 30, 50, and 100 kWh, the corresponding effective charging rates are 30, 50, and 50, respectively.

\myparagraph{Charging time calculation based on effective charging rate}
If the effective charging rate of a request r at a charging station c is $CR(c,r)$, $t^-_{c}$ and $t^+_{c}$ define the time upon arrival and departure from the station $c$, and $SoC^-_{c}$ and $SoC^+_{c}$ represent the state of charge of the vehicle at that time respectively, then the charging time is calculated as, 

\vspace{-5pt}
\begin{equation}
    ct(c, r) = 
      \frac{(SoC^+_{c} - SoC^-_{c}) \times B_r}{CR(c,r)}  
\label{eq:charge}
\end{equation}
\vspace{-5pt}

Note that, EVs usually follow a linear charging time for charge up to 80\%, and a slower charging rate for higher percentage of charge~\cite{schoenberg2022reducing}. Our experiments have shown that over 90\% of the optimal paths are formed by partial recharging, each less than 80\% of charge. Therefore, we consider that the charging time is linear for simplicity in this paper. A non-linear function to calculate charging time instead of Eqn.~\ref{eq:charge} can directly be applied to our proposed algorithm, and the experiment results will still be almost similar. Lastly, the wait time at a CS~$c$ is calculated as, $wt(c, r) =(t^+_{c} - t^-_{c}) - ct(c, r)$.

Our goal is to find OCP with partial recharging for each EV request $r$, such that the route is not only optimal for the current request, but also optimal for all the requests in the stream as much as possible. Before we present the details of such optimization while ensuring each time the optimality of the route for the current request is preserved, we first present how we have modelled the problem in a graph, provide necessary definitions, and then formally formulate our problem.

\section{Modeling and formulating the problem}\label{sec:model}

In this section, we first present how we have modelled the EV route planning problem in a graph. We modelled both the spatial and temporal aspects (waiting time and charging time) of the problem within the graph itself. \textbf{The novelty and key reason for modelling in this way is}, this graph provides a general model for the EV charging problem where it is straightforward to incorporate several different variations of the problem as different graph parameters, which is of independent interest (details in Section~\ref{Discussion}). Specifically, (i) We propose a graph with time-dependent self-loops to model the problem, to efficiently capture the timeslot information within the graph itself. (ii) We specify the problem statement on this new setting. (iii) We show that the problem is NP-hard.

\vspace{-5pt}
\subsection{Routing network with charging stations as a Graph with Time Dependent Self-loops (GTDS)} Consider a road network is given, where the vertices are the road intersections and the edges represent the roads. On such a road network, we incorporate the charging stations, additional edges, and some other parameters to create a `Routing network' for our EV charging problem as a Graph with Time Dependent Self-loops (GTDS), $G = (V, E, W, U, L)$, where
\begin{enumerate}
    \item $V$ is a set of vertices where a vertex $v \in V$ is either a road intersection or a charging station.
    \item There is an edge $e_{v_i,v_j} \in E, i\neq j$ from vertex $v_i$ to $v_j$, representing the road connecting $v_i$ and $v_j$. 
    \item Each edge $e_{v_i,v_j}$ is associated with a weight $w(e)$, representing the travel time from $v_i$ to $v_j$ ($i\neq j$).
    \item There is a second type of edge associated with each vertex representing a CS. For each CS vertex $c$, $l_{c} \in L$ is a self-loop edge for $c$ (details in Section~\ref{subsec:selfloop}).
    \item Each edge is also associated with $u(e)$, denoting the amount of resource unit (i.e., charge) gained or lost during traversal. For edge $e \in E$ connecting two vertices, $u(e)$ is negative, representing the quantity of charge lost while traversing the edge. If the edge is $l_c \in L$ at a charging station $c$, then $u(l_c) \geq 0$, representing the charge gained at $c$. The value of $u(l_c)$ is time-dependent and may vary at each timeslot (more details below). 
\end{enumerate}

Note that, while $u(e)$ can be pre-computed for EVs with a certain energy consumption rate, the value $u(e)$ can simply be adjusted in runtime for any EV with a different energy consumption rate by multiplying $u(e)$ with the ratio of the energy consumption rates.

\vspace{-10pt}
\subsection{Time-dependent self-loops to capture temporal information}\label{subsec:selfloop} As mentioned above, each vertex representing a CS is associated with a self-loop edge. We now present how such edges and an additional data structure,  `time-dependent resource table (TRT)' can capture the temporal information.

\myparagraph{Time-dependent resource table (TRT)} In this table, each column corresponds to a CS, and the information is organized for the next $n$ timeslots (including the current timeslot) in a sliding window fashion. The purpose of this table is to track which CS is reserved for specific timeslot(s), and if a CS is available at a timeslot, then how much charge can be obtained by a vehicle from there. Here, \textbf{(1)} If a particular timeslot is already reserved at a CS, then the corresponding value is 0 in TRT, denoting that no charge can be gained from that station at those timeslots by another EV. \textbf{(2)} If a particular timeslot is available at the charging station~$c$, then the resource gained by a request $r$ at $c$ at each such timeslot can be calculated based on the effective charging rate $CR(r,c)$ and the length $t$ of the timeslot in minutes (e.g,. 5 minutes) as, $CR(r,c) \times t/60$.

\begin{figure}[!tbh]
\centering
 \includegraphics[width=\linewidth]{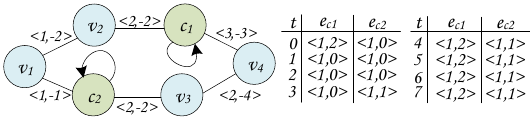}\vspace{-8pt}
  \caption{Example of a GTDS where each $v_i$ represents vertices, $c_1$ and $c_2$ represent two CSs. The weight $w(e)$ and resource unit $u(e)$ of each edge is shown as a pair of $<w,u>$ values with each edge. The shown table is the TRT, with a column for each self-loop for 8 timeslots. Each pair $<1,u>$ means that, by spending a timeslot, EV can gain $u$ unit of charge. If a timeslot is reserved at a CS, $u$ is zero for that entry (e.g., the first 3 timeslots in $c_2$).}
\label{fig:GTDSExample}
\vspace{-10pt}
\end{figure}

\begin{example}[GTDS and TRT]
Figure~\ref{fig:GTDSExample} shows a GTDS with two CSs. If the charging rate of $c_1$ and $c_2$ is $36$ and $12$ kWh, then for an EV with a $\Gamma$ of $24$ kWh, the effective charging rate in $c_1$ and $c_2$ will be $min(36, 24) = 24$ and $min(12, 24) = 12$, respectively. If timeslots take $5$ mins, the EV can get 2 kW from $c_1$ and 1 kW from $c_2$ at each timeslot. These values are shown in the TRT of Figure~\ref{fig:GTDSExample} (details in the figure caption). 
\end{example}

\vspace{-12pt}

\begin{example}[Computing total time of a specific path] For the given path $p = \{v_1, v_2, c_1, v_4\}$ in the GTDS shown in Figure~\ref{fig:GTDSExample}, let's calculate the total travel time for an EV with a maximum battery capacity of $B=5$. The EV commences the trip at $t=0$ with $B=5$ and reaches $v_2$ at $t=1$, utilizing $2$ out of its $5$ resource units. Then it arrives at $c_1$ at $t=3$ with only $1$ remained resource. At $t=3$, $e_{c_1}$ is $<1,0>$, signifying no gain or loss of charge while traversing $e_{c_1}$ at $t=3$. Subsequently, at $t=4$, 2 resources are acquired and the battery is $3$. The EV can leave $c_1$ at $t=5$ and finally reaches $v_4$ at $t=8$. The total travel time of this path is $8$.

\end{example}

\vspace{-12pt}
\subsection{Problem formulation}
In this section, we first formally formulate our problem, and then show that the problem is NP-hard. 

\myparagraph{Problem statement} Given a stream of requests $R$, our \newprob problem is to find a path for each $r \in R$ such that (i) the path is guaranteed to be an $OCP(r)$ for the current request $r$, and (ii) the total travel time for all requests in $R$ is minimized. 

\myparagraph{NP-hardness} We show that even just finding the $OCP(r)$ of an EV request $r$ on a GTDS is NP-hard by reduction from the constrained shortest path (CSP) problem.

\begin{lemma}
\label{lemma:NP}
Finding $OCP(r)$ is an NP-hard problem.
\end{lemma}

\begin{proof}
Let $G =(V, E, W, U)$ be a graph, where $V$ represents vertices, $E$ denotes edges, $W$ is the set of edge weights, and $U$ signifies the resources needed to traverse the edges. Given $o \in V$ as the origin, $d \in V$ as the destination, and $M$ denoting the maximum available resource, the classical CSP from vertex $o$ to vertex $d$ is the path with the minimum total weight while ensuring that the total used resources along the path do not exceed $M$. The CSP is an NP-hard problem~\cite{garey1997computers}.

Consider a generic case of the CSP problem involving an additional edge type $L$, where the value associated with each edge in $L$ represents the resource gained by traversing it. Consequently, solving the CSP problem is equal to finding a path with the minimum total travel time from $o$ to $d$, while ensuring that the total used resources along the path do not exceed $SoC$, and the resource needed to travel along an edge is equal to the energy required $u$ to traverse that edge. Again, finding $OCP(r)$ from $o$ to $d$ where the maximum available resource $M$ corresponds to $SoC$, 
the edge travel times corresponding to weights $W$, and the energy for traversing an edge correspond to $U$, is equivalent to solving the CSP problem. So, we have shown that a solution to the CSP problem exists iff there exists a solution to the generic case of finding $OCP(r)$, where the CSP is reducible to finding the $OCP(r)$ in polynomial time. Therefore, finding $OCP(r)$ is an NP-hard problem.
\end{proof}

As explained in Section~\ref{subsec:other_related}, while several approaches have been proposed to address the CSP problem, most of them cannot be extended to the \newprob as they are confined to either positive or negative edge weights~\cite{morlock2019time}, but not both, which does not align with the \newprob problem. Even solutions applicable to CSP with both positive and negative edges lack in finding the optimal route for an EV while considering future EVs and offering a proactive solution. 

\section{Proactive EV route planning algorithm}\label{ProactiveAlgorithm}

The key idea of our approach is, there can be many optimal charging paths for a request with the same minimum travel time. Hence, The \newprob seeks to find such an OCP(r) for the current request $r$ in the stream, which excludes the CSs and/or the timeslots that are expected to be in the OCP of the future requests, as much as possible. In the following, we present our solution with the details of the steps.

\subsection{Pre-processing} In this section, we present the pre-processing steps that we propose to improve the efficiency of our NP-hard \newprob problem for stream of EV requests. 

\myparagraph{Reduction of the search space to charging network only} Given the fixed locations of CSs and the expectation for users to choose the fastest path between CSs during long trips, we first reduce the routing network to a graph with CSs only. This strategic pre-processing step eliminates redundant computations between fixed locations, enhancing overall solution efficiency. The resultant CS-only graph is constructed with vertices representing CSs and edges connecting CS pairs within a specified threshold, i.e., maximum range of EVs. The edges maintain their weight $w(e)$, denoting travel time of an edge $e$ between the pair of CSs, and resource changes during traversal are captured by $u(e)$. Self-loops $L$ at CSs remain unchanged. Note that the values associated with an edge in this graph can easily be updated in real time to incorporate any changes in travel times and energy consumption. Throughout this paper, we refer to this simplified graph, focused exclusively on CSs without road intersections, as the routing network. 

\myparagraph{Landmarks to substitute source and destination} When a request $r$ arrives, one approach is to add its origin and destination as vertices in the graph~\cite{schoenberg2022reducing}, create the necessary edges, and compute their weight $w(e)$ and resource $u(e)$ values. However, to further reduce runtime computations in this NP-hard problem, we introduce `landmark' vertices. Landmarks are selected from the routing graph and are connected to CSs within the maximum range of EVs. The weight $w(e)$ and the resource $u(e)$ of each such edge $e$ are calculated and included in the routing network. Subsequently, when a request $r$ is received, we replace its origin and destination locations with the nearest landmark vertices. This substitution renders other landmarks irrelevant for the OCP calculation, allowing for the inclusion of a larger number of landmarks in the routing network without affecting subsequent computations.

\myparagraph{Future request predictor} As we mentioned, given a stream of requests, our objective is to find an optimal charging path for each request $r_i$, while also minimizing the total travel time for the other requests in the stream. 
While finding the OCP for $r_i$, we are unaware of the upcoming requests $\{ \forall r_j | t_{r_j} > t_{r_i}\}$ due to the streaming nature of requests. To tackle the \newprob, we propose a solution involving a request predictor, treated as a black box for finding a proactive optimal solution.

Our paper focuses on utilizing future demand information, rather than predicting it directly, to plan optimal charging paths. For simplicity, we assume that upon each request arrival, a set $R'$ of future $n$ requests is provided, each containing the relevent information similar to any EV routing request. It is also feasible to adapt the existing spatial-temporal prediction method~\cite{yao2018deep} for predicting such future requests.

Importantly, our approach maintains the optimality of the charging path solution for the current request regardless of the accuracy of the future request predictor. This means that even if all $n$ future requests are predicted incorrectly, the solution for the current request remains guaranteed to be optimal.

\subsection{Influence factor} In the following, we first propose some key concepts of the influence of one charging path on another, which will be later used to present the steps of our approach in details.

\myparagraph{Correlation of two paths} In a GTDS $G$, two optimal charging paths $p$ and $p’$ of two different EVs exhibit correlation iff there exists a time $t$ where both paths use the same self-loop edge in $G$ (i.e., a specific timeslot at a particular CS is part of the optimal charging paths of both EVs). If function $co(p,p’)$ defines the correlation between $p$ and $p’$, then it results in 1 if there is a correlation between $p$ and $p’$, otherwise 0.

\begin{example} Consider the following charging paths (Superscripts denote timeslots, and subscripts denote CS or landmark): $p_1 = \{ l_1^0, c_1^{10}, c_1^{11}, c_1^{12}, c_1^{13}, c_1^{14},l_2^{26}\}$ , $p_2 =\{l_2^{1}, c_2^{10},c_2^{11}\\, c_2^{12}, c_2^{13}, l_1^{26}\}$, $p_3=\{l_2^{1}, c_1^{13}, c_1^{14}, c_1^{15}, c_1^{16}, l_1^{26}\}$, $p_4 = \{l_3^{4}, c_1^{15},\\ c_1^{16}, c_1^{17}, c_1^{18}, l_1^{28}\}$, where $p_1 \in OCP(r_1)$, $p_2,p_3 \in OCP(r_2)$, and $p_4 \in OCP(r_3)$. No correlation exists between $p_1$ and $p_2$ since they are charged at different CSs. The same holds true for $p_1$ and $p_4$ as their charging times do not overlap, despite using the same CS. Conversely, $p_1$ and $p_3$ exhibit correlation because these paths consist of the same CS ($c_1$) at the same timeslots ($c_1^{13}, c_1^{14}$). So, the values of $co(p_1, p_2)$, $co(p_1, p_3)$, and $co(p_1, p_4)$ are 0, 1, and 0, respectively.

\end{example}

The correlation of a path $p$ indicates which paths for the future requests become unavailable if $p$ is chosen for the current request. The unavailability of a specific path of the upcoming requests can result in an increase in their travel time. Therefore, it is essential to establish a factor that quantifies the influence of a path on future requests.

\myparagraph{Influence factor of a charging path on a set of requests} 
The influence of charging path $p \in OCP(r)$ on a set of requests $R’ = \{ r’ \in R | t_{r’} > t_r \}$ can be characterized from two perspectives: i) \textbf{Direct influence:} Since our goal is to minimize travel times for all vehicles, we denote the direct impact of a path $p$ on a set of requests $R’$ as the increased travel time that future requests $R’$ will experience as a result of their correlation with $p$. ii) \textbf{Indirect influence:} Although the correlation with path $p$ does not always result in extended travel times for $R'$, due to the presence of multiple optimal charging paths with the same minimum travel time, it may lead to a decrease in the number of available optimal paths for each request in $R’$. The likelihood of experiencing longer travel times for future requests is higher when there are fewer optimal paths with the minimum travel time available. So we define indirect influence of path $p$ on $R’$ as the number of optimal paths for $R'$ which has a correlation with path $p$.

\myparagraph{Proactive-optimal charging path} Given a set of optimal charging path $OCP(r)$ and a predicted upcoming requests set $R'$, the charging path $p_{po} \in OCP(r)$ is referred to as proactive-optimal charging path when $p_{op}$ exhibits the lowest direct influence factor. If there are multiple paths with the same direct influence, the path with the lowest indirect influence factor among them is considered as $p_{op}$.

Our proposed solution to find a proactive optimal charging path for each $r \in R$ contains two components: an algorithm to find all $OCP(r)$ of the current request $r$, and calculating influence factor for each charging path in $OCP(r)$. We propose an efficient two-phase algorithm to find all $OCP(r)$ of a request $r$ in Section~\ref{TwoPhaseAlg}, and an effective refinement-based approach to calculate influence factor in Section~\ref{InfluenceFactorAlg}.

\subsection{Finding OCP of a request}\label{TwoPhaseAlg}
As mentioned in Section IV, the problem of finding OCP for one request is an NP-hard problem. Hence, we propose an efficient solution to calculate the travel time of paths that include charging and waiting time using TRT (presented in Section IV), and a routing table (explained later). In the initial phase, we proceed forward from the source point and compute the various potential departure times for each CS based on a specific SoC. Subsequently, these values are stored in the routing table and the process persists until the destination is reached. Upon arrival at the destination and obtaining an upper bound for the length of a feasible route, we commence our second phase, which is reverse traversal from the destination to the source. Employing the information stored in the routing table, various alternative paths are constructed while considering the latest permissible departure time from each CS. Before explaining the algorithm, we define `Routing table'. 

\myparagraph{Routing table} The routing table serves as a straightforward mechanism for storing arrival and leaving information for each CS during the forward phase. Each row of the table stores the following information: the arrived node, arrival time and SoC, the ancestor node (i.e., immediate predecessor to the arrived node in the current path), charging time, departure SoC, and departure time from the CS. Now we present our algorithm.

\subsubsection{\textbf{Forward Pass}}

Each EV possesses SoC and requires recharging, which can take a different amount of time considering the charging rate and status of a CS (i.e., reserved or available) at different times. Thus, \textbf{arriving earlier may not necessarily lead to departing from the CS earlier.} Moreover, the path with the shortest travel time can consist of multiple partial recharging. So, we propose our efficient techniques that utilize the TRT and routing table to find the OCP. 

\myparagraph{Leaving Charging Bucket} 
In our algorithm, each vehicle is capable of departing from any CS with a SoC that corresponds to a predefined value referred to as a Leaving Charging Bucket ($LCB$). For example, $LCB = \{50, 100\}$ means each EV can leave the CS with SoC equal to either 50 or 100 percent. 

\begin{figure}[t]
\centering
 \includegraphics[width=\linewidth]{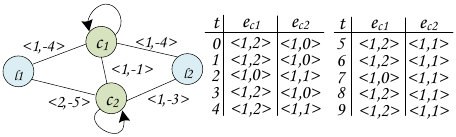}
 \vspace{-20pt}
\caption{GTDS of Example 4. $l_1$ and $l_2$ are the source and destination, $c_1$ and $c_2$ are CSs with effective charging rate $2$ and $1$ per timeslot, respectively. }
\label{fig:RoutingAlgorithmExample}
\end{figure}

\begin{table}[h]
\centering
\vspace{-10pt}
\caption{Routing table of Example 4}
\vspace{-5pt}
\begin{tabular}{c | c   c  c  c  c  c   c c } 
 \hline
 Rows & V & anc & $t^a$ & $SoC^{anc}$& $SoC^a$ & $SoC^d$ & $t^{ch}$ & $t^d$ \\ [0.5ex] 
 \hline
 1& $l_1$& - & - & - & - & 5 & - & 0\\ 
 2 &$c_1$ & $l_1$&  1 & 5 & 1 & 3& 1 & 2\\
 3& $c_1$& $l_1$ & 1& 5 &1 & 5& 2 & 5 \\ 
 4 &$c_2$ & $l_1$& 2& 5& 0 & 3& 3 &7\\
 5 &$c_2$ &$l_1$ & 2& 5&0 & 5& 5 & 9\\
 6& $c_2$ & $c_1$& 3& 3 & 2& 3& 1 & 5\\
 7& $c_2$&$c_1$ &3& 3 &2 & 5& 3 & 7\\ 
 8& $c_2$ & $c_1$& 5& 5 & 4& 5& 1 & 6\\
 9 &$l_2$& $c_1$ & 6& 5& 1 & -& - & 6\\
 10 &$l_2$& $c_2$ & 6& 3& 0 & -& - &6\\
 \hline
\end{tabular}
 \vspace{-10pt}
\label{table:RoutingTable}
\end{table}

\myparagraph{Algorithm} Our algorithm employs a best-first search over $GTDS$, utilizing a list to track nodes (vertices in $GTDS$) and their associated SoC for expansion. In each iteration, the algorithm selects the node-SoC pair with the lowest time for expansion and identifies reachable neighbor nodes at the current SoC level. Instead of considering arrival time, the algorithm calculates \textbf{the earliest time to reach a specified SoC level} in each neighbor node and stores it in the list. The routing table captures all details regarding arrivals, leave times, and leave SoC. Additionally, the algorithm maintains an upper bound based on the best-travel-time path found so far from the source to the destination, aiding in pruning paths that cannot surpass this upper bound. Algorithm~\ref{alg:ForwardShortestPath} describes the proposed algorithm in details. The steps are explained as follows:

\begin{itemize}
    \item Initialization: The origin node with the initial $SoC$ is added to the open list $\Lb$ (Line 1). 
    \item The iteration over $\Lb$ persists until all items are explored. Every time, the item with the minimum time is selected and the node-SoC pair is marked as visited (Line 3). 
    \item Using the $GTDS$, all reachable neighbor nodes based on the available $SoC$ is determined (Line 4). The arrival time for each reachable node $n$ from the current node $m$ is the summation of the departure time from $m$ and the driving time from $n$ to $m$ (Line 5). The remaining charge upon arrival at $m$, denoted by $SoC^a$, is calculated by subtracting the required charge for travel from $n$ to $m$ with the $SoC$ at the departure time from $n$ (Line 6). 
    \item Records in routing table: For charge value in $LCB$, we find the first possible consecutive timeslots that the vehicle can get charged (Lines 7 to 14). So, the earliest leave time for node $n$, given its arrival time $t^a$ and charge $SoC^a$ for each bucket in $LCB$, is considered as $t^d$ for that node. We record ancestor node ($anc$), arrival time from the ancestor($t^a$), leaving $SoC$ from ancestor ($SoC^{anc}$), departure state of charge ($SoC^d$) and time ($t^d$) in the routing table (Line 17).
    \item Upper bound computation: When we reach the destination for the first time, we set the arrival time as upper bound $ub$, 
    indicating the existence of a charging path with a travel time of at least $ub$, and from now on any path with a longer time will be rejected. If the neighboring node $n$ is the destination and the travel time is less than the current $ub$, we update the upper bound (Lines 18-19). 
    \item If $n$ is the destination and the travel time is not less than the current $ub$, we check if $n$ with SoC $b$ is visited. If not, we include $n$ along with the current charge and departure time in $\Lb$ for later exploration (Lines 20-21). If $n$ is in $\Lb$ but has not been explored, we check whether the current solution provides a better travel time and update $\Lb$ accordingly (Lines 22-23). 
\end{itemize}

\begin{algorithm}[t]
\footnotesize
\DontPrintSemicolon
  
  \KwInput{GTDS $G$, EVRoutingRequest $r$}
  \KwOutput{Routing table and optimal path upper bound}
  
  $\Lb \gets (r.origin, r.SoC, r.time)$, $\Lt \gets null$, $ub \gets \infty$, $\Lv \gets null$ \;
  
  \While{$\Lb$ has items}{
  $(m, s_m, t_m) \gets$ Sort $\Lb$ by $time$ and Pop
  
  $\Lv.add((m,s_m))$
  
  \ForEach{reachable neighbor $n$ from $m$}
    {
        
        {$t^a = t^d \gets t_m + w_{m,n}$}
        
        $SoC^a = SoC^d \gets s_m - (r_{m,n}/B_r)$
        
        \ForEach{$b \in LCB$}{
            $t^w \gets 0, t^{ch} \gets 0$
            
            \While{$SoC^d < b$}{
                \If{$r_{c,t^d}$ = 0}{
                    $SoC^d = SoC^a, t^{ch} \gets 0$
                }
                \Else{
                $SoC^d += r_{c,t^d}, \: \: t^d += w_{c,t^d}$
                
                $t^{ch} +=w_{c,t^d}$
                }
            }
        
        \If{$t^d > ub $}{
        Go to line 2
        }
        Add $(n, m, t^a, c.SoC, SoC^{a}, b, t^d)$ to $\Lt$
        
        \If{$n = r.dest$ and $ub <t^d$}{
        $ub \gets t^d$
        }
        
        \If{$(n, b)$ not in $\Lv$}{
        $\Lb.add(n, b, t^d)$
        }
        
        \If{$t^d < $ leave time from $n$ with $b$ in $\Lb$}{       
        Update $\Lb$ with $(n, b, t^d)$}
     }   
    }
    } 
  \Return $\Lt, ub$ \;
\caption{ForwardPass($G$, $r$)}\label{alg:ForwardShortestPath}
\end{algorithm}

\myparagraph{Complexity} If $\eta_{min}$ represents the minimum charging rate of CSs, the computation of charge time is in the order of $O(B_r/\eta_{min})$. Since this calculation is performed for each pair of edges and LCB, the worst-case complexity of the forward path algorithm is $O(|E| \cdot |LCB| \cdot (B_r/\eta_{min}))$.

\begin{example} In Figure~\ref{fig:RoutingAlgorithmExample}, we want to find OCPs from $l_1$ to $l_2$ for an EV with $B_r=5$ and $LCB=\{3, 5\}$, starting at $t=0$. We start from $l_1$ and find all reachable neighbors, $c_1$ and $c_2$. The arrival time ($t^a$) and state of charge ($SoC^a$) upon reaching $c_1$ are 1 and 1, and for $c_2$ are 2 and 0, respectively. Considering $c_1$, with $LCB = \{3,5\}$, we need 2 or 4 units of charge start from $t=1$ to leave $c_1$. As $e_{c_1}$ at $t=1$ is $<1,2>$, i.e., spending one timeslot results in $SoC^d=3$. Therefore, $(c_1,3,2)$ is added to $\Lb$, which means there is a path from source to $c_1$ where we can leave $c_1$ at $t^d=2$ with $SoC^d=3$. The routing table is shown in Table~\ref{table:RoutingTable}. The first row indicates the starting node, starting $SoC^d=5$ and departure time $t^d=0$. The second row is for $(c_1,3,2)$. As $e_{c_1}$ at $t=2$ is $<1,0>$, this timeslot is already reserved. So, $t=3$ and $t=4$ will be chosen. For $SoC=5$, the departure time will be 5. Hence, $(c_1,5,5)$ is added to $\Lb$ and the corresponding values are shown in row 3 of Table~\ref{table:RoutingTable}. This procedure is also done for $c_2$, where $(c_2,3,7)$ and $(c_2,5,9)$ are added to $\Lb$, and rows 4, 5 are added in routing table.

Next, we choose the option from $\Lb$ with the lowest travel time, which is $(c_1, 3, 2)$. Traversing from $c_1$ to $l_2$ is not possible, as the required energy is 4. The only reachable neighbor is $c_2$. So, it is possible to reach $c_2$ again, this time from $c_1$ with $SoC^a=2$, which entails to have a leave times from node $c_2$ at 5 and 7, respectively (rows 6 and 7). As we have found a faster departure time from node $c_2$, the previous values in $\Lb$ is updated from $(c_2,3,7)$ and $(c_2,5,9)$ to $(c_2,3,5)$ and $(c_2,5,7)$, respectively. Then, the subsequent iteration selects $(c_1,5,5)$. By exploring $c_1$ with $SoC^d=5$ at $t=5$, we find a path to destination $l_2$ at $t=6$ with $SoC^a=1$ (row 9). So, $ub$ is set to 6. The next option is $(c_2,3,5)$. We find another path to $l_2$ reaching at $t=6$ from $c_2$ (row 10). Consequently, all other options from $\Lb$ are discarded as their departure time is larger than 6. 
\end{example}

\subsubsection{\textbf{Backward Pass}}

The purpose of this step is to generate all possible paths utilizing the routing table generated in the previous step. To accomplish this objective, we initiate reverse exploration from the destination, considering the upper bound using the algorithm presented in Algorithm~\ref{alg:BackwardShortestPath}. 

\myparagraph{Algorithm} Initially we create an open list $\Lb$ and add the destination and $ub$ (found at the end of forward pass) in it (Line 1). In each iteration, an item is selected from $\Lb$ representing the current node $n$, departure SoC $s_n$, departure upper bound $ub_n$, and the constructed path up to that point (Line 3). Then, Rows from the routing table where departure from $n$ occurred within the specified upper bound ($t^d \leq ub$) and $SoC^d$ matches the determined value are retrieved (Line 4). Each row corresponds to a potential distinct path. For each retrieved row, the upper bound of the ancestor node is calculated by subtracting the driving time from the ancestor to $n$ and the charging time at $n$ (Lines 5 to 8). Note that, the charge time is zero when $n$ is either the origin or the destination. If there is waiting time at the current node, this wait time is considered as a late arrival at $n$. An ancestor node capable of reaching $n$ at a later time than other nodes, while still allowing the vehicle to depart within the upper bound and meeting the required SoC level, is deemed a viable route and added to the list of paths for exploration. If the ancestor is the origin of $r$, the generated path is designated as a charging path (Lines 9-10); otherwise, exploration continues through all open nodes until reaching the origin.

\begin{algorithm}[h]
\caption{BackwardPass($G$, $r$, $\Lt$, $ub$)} \label{alg:BackwardShortestPath}
\footnotesize
\DontPrintSemicolon
  \KwInput{GTDS $G$, EVRoutingRequest $r$, RoutingTable $\Lt$, UpperBound $ub$}
  \KwOutput{All charging paths with minimum travel time}
  \SetKwBlock{Beginn}{beginn}{ende}
  $\Lcp \gets null$, $ \Lb \gets (n=r.dest, soc=*, t=ub, p=[(r.dest,*)])$

  \While{$\Lb$ has items}{
  {$(n, s_n,ub_n ,p_n) \gets$ Pop first item from $\Lb$
  
  $Rows \gets$ All rows from $\Lt$ where $V = n$ \& $ t^a <= ub_n$ \& $SoC^d=s_n$ 
  }
  
  \ForEach{$r \in Rows$}{
  
  $m = r.anc, s_m = r.SoC^{anc}$
  
  $ub_m = ub_n - w_{m,n} - r.t^{ch}$
  
  $p_m \gets p_n, p_m.add(m, s_m)$
  
  \If{$m = r.origin$}{
  $\Lcp.add(p_m)$
  }
  \Else{
  $\Lb.add(m, s_m, ub_m, p_m)$
  }
  }
  }
  \Return $\Lcp$ \;
\end{algorithm}

\myparagraph{Complexity} If $E_{max}$ defines the maximum number of edges satisfying the upper-bound constraint for a specific node, then the worst-case complexity of the backward algorithm is $O(|V|.|E_{max}|)$.

\begin{example} Here, we demonstrate the backward phase using Table~\ref{table:RoutingTable}. Initially, we put $(l_2, *, 6, [])$ to $\Lb$. This entry shows that we are looking for any row in the Table~\ref{table:RoutingTable} where $V=l_2$, $t^d = 6$. The value matches with rows 9 and 10. 
First consider row 9 with ancestor $c_1$. Since $l_2$ is the destination, we deduct the driving time from $c_1$ to $l_2$ from the current $ub$ and add ($c_1, 5, 5, [(l_2,*)]$) to $\Lb$. Similarly, for row 10, ($c_2, 3, 5, [(l_2,*)]$) is added to $\Lb$. By choosing the next item from $\Lb$, we retrieve all rows from the routing table that satisfy $v=c_1$, $SoC^d=5$, and $t^d\leq5$ (the third row). As $c_1$ is a CS, we subtract the charge and drive time between $c_1$ and its ancestor. Therefore, considering a charge time of 2, the ancestor's $ub$ will be $5-2-1$, and SoC will be 5. Consequently, a new item is added to $\Lb$ as $(l_1, 5, 2, [(l_2,*), (c_1, 5)])$. The procedure is carried out for $(c_2, 3, 5, [(l_2,*)])$, and $(c_1, 3, 3, [(l_2,*), (c_2,3)])$ is added to $\Lb$. For $(l_1, 5, 2, [(l_2,*), (c_1, 5)])$, as $l_1$ is the source, the path $\{(l_1,5), (c_1, 5), l_2\}$ is one of the optimal paths. By exploring the other item $(c_1, 3, 3, [(l_2,*), (c_2,3)])$ in $\Lb$, we reach to 
$(l_1, 5, 1, [(l_2,*), (c_2,3), (c_1,3)])$, which leads to the source. So, the second optimal path will be $\{(l_1,5), (c_1,3), (c_2, 3), l_2\}$. The travel time of both paths is 6.
\end{example}

\vspace{-15pt}
\subsection{Influence factor calculation}\label{InfluenceFactorAlg}

The proposed influence factor for a set of optimal paths $OCP(r)$ depends on the optimal paths of every predicted future requests. So, we utilize the two-phase algorithm again, but this time for calculating the influence factor. As not all OCPs of the current request have influence on all future requests, we first propose a refinement step to determine the future requests for which calculating influence is not necessary.

\myparagraph{Refining the future requests}
The purpose of the refinement algorithm is to limit exploration of the search space for requests that exhibit no correlation with the optimal paths for request $r$. For this purpose, we propose `Latest Leave Time To Impact' ($LLTI$) in Algorithm~\ref{alg:LatestLeaveTimeToInfluence}. Initially, the leave time from any CS for any $OCP(r)$ is recorded in the $LLTI$ (Lines 2-5). Subsequently, for every node in the $GTDS$, we compute the upper bound $ub_m$ for node $m$ as a neighbor of $n$ to demonstrate that a vehicle must depart from $m$ no later than $ub_m$ to potentially affect one of the paths in $OCP(r)$ (Lines 8 to 11). This condition is later used to refine future requests, where calculating influence is unnecessary.

\begin{algorithm}[h]
\footnotesize
\DontPrintSemicolon
  \KwInput{OCP $P$, GTDS $G$}
  \KwOutput{Latest Leave time to Influence on $P$}
  $\Ln \gets $all nodes in $G$, $LLTI[\Ln] \gets -1$

  \ForEach{$p \in P$}{
  \ForEach{$s \in p$}{
  \If {$LLTI[s] < $ latest charge time at $s$}{
      $LLTI[s] = $ latest charge time at  $s$}
  }
  }
  \While{$\Ln$ has items}{
  $n \gets$ pick from $\Ln$ with highest $LLTI$, $\Ln.remove(n)$
  
  \ForEach{neighbor $m$ from $n$}{
  $ub_m \gets LLTI[n] - w_{n,m}$
  
  \If {$LLTI[m] < ub_m$}{
      $LLTI[m] = ub_m$
  }}}
  \Return $LLTI$
  \caption{LeaveTimeToImpact($P$, $G$)}
  \label{alg:LatestLeaveTimeToInfluence}
\end{algorithm}

\myparagraph{Algorithm for influence factor calculation}
To calculate influence, we need to calculate $OCP(r’)$ for all requests $r’ \in R’$. However, by utilizing LLTI, we can determine if further exploration of the search space to find the optimal path for $r’$ can generate paths that are correlated with any of the paths $p \in OCP(r)$. If it is not possible to create a correlated path with OCP(r), we will stop exploring the search space for $r’$. 

Since $OCP(r’)$ generates all optimal paths with the minimum travel time, it is guaranteed that all the paths will have the same travel time. This makes it impossible to calculate the direct influence. To overcome this, we introduce a threshold, denoted as $\epsilon$, to establish an upper bound limit for the travel time of paths included in $OCP(r’)$. In Algorithm~\ref{alg:InfluenceCalculator}, we compute all paths with a length equal to or less than the shortest possible travel time plus $\epsilon$ for each $r’ \in R’$ (Lines 3-13). The feasibility of generating a path to impact $OCP(r)$ is examined (Line 6), and the upper bound is set as the sum of the travel length of the shortest found path plus $\epsilon$ (Lines 11-12). After generating all paths using the backward pass in Line 14, we determine the direct influence of path $p$ on future requests in Lines 16-19 by calculating the amount of increase in the travel time of future requests $R'$ for each path $p \in P_r$. If path $p$ renders all paths for request $r' \in R'$ unavailable, then we set the $ub$ for request $r'$ as travel time for that request (Line 18). Finally, the indirect influence in Lines 20-21 is calculated as the number of paths that become unavailable because of $p$.

\subsection{Proactive route planning algorithm}

The Algorithm~\ref{alg:ProactiveRoutePlanning} presents the overall proactive route planner. First, all optimal paths for request $r$ are identified in Lines 1-2. Then, the influence factor for each path is calculated (Line 3). The proactive path for $r$ is then determined by selecting the path with the lowest direct influence. If there are multiple paths with the same minimum direct influence, the proactive path is chosen based on the minimum indirect influence.  

\begin{algorithm}[t]
\footnotesize
\DontPrintSemicolon
  
  \KwInput{GTDS $G$, OCP $P_r$, RoutingRequests $R'$, Threshold $\epsilon$}
  \KwOutput{All charging paths with minimum travel time}

   $LLTI \gets LeaveTimeToImpact(P_r, G)$   

   $DirectInf[P_r] \gets 0$, \: $IndirectInf[P_r] \gets 0$
   
   \ForEach{$r' \in R'$}
   {
  $\Lb \gets (r'.origin, r'.SoC, r'.time)$, $\Lt = \Lv \gets null$, $ub \gets \infty$
  
  \While{$\Lb$ has items}{
  \If{$t_m <= LLTI[m] \: \: \forall (m, s_m, t_m) \in \Lb $}
{
  Lines 4 - 16 of ForwardPath Algorithm
  
 \If{$t^d > ub + \epsilon $}{
        Go to line 3
        }
        Add $(n, m, t^a, c.SoC, SoC^{a}, b, t^d)$ to $\Lt$
        
        \If{$n = r.dest$ and $ub <t^d + \epsilon$}{
        $ub \gets t^d + \epsilon$
        }
    Lines 22 - 25 of ForwardPass algorithm
    } 
    }
    $P_{r'} = BackwardPass(G, r', \Lt, ub)$
    
    \ForEach{$p \in P_r$}{
    
    $P^{nco}_{r'}$ = paths from $P_{r'}$ where $co(p_{r'},p) = 0$
    
    $t_{p_{r'}}$ = min travel time in $P_{r'}$
    
    $t'_{p_{r'}}$ =  min travel time in $P^{nco}_{r'}$, set $ub_{r'}$ if $P^{nco}_{r'}$ is $Null$

    $DirectInf[p] += t'_{p_{r'}} -  t_{p_{r'}}$ 

    $P^{co}_{r'}$ = paths from $P_{r'}$ where $co(p_{r'},p) = 1$
    
    $IndirectInf[p] += len(P^{co}_{r'})$ 
    }
    }
  \Return $DirectInf$, $IndirectInf$ \;
\caption{InfluenceFactorCalculator($G$, $P_r$, $R'$, $\epsilon$)}\label{alg:InfluenceCalculator}
\end{algorithm}

\begin{algorithm}
\footnotesize
\DontPrintSemicolon
  \KwInput{EVRoutingRequest $r$, RoutingRequests $R'$, GTDS $G$}
  \KwOutput{A proactive path}
  
  $\Lt, ub = ForwardPass(G, r)$ 
  
  $P_r = BackwardPass(G, r, \Lt, ub)$
  
  $Inf_d, Inf_i = InfluenceFactorCalculator(G, P_r, R', \epsilon)$
  
   $idx = argmin(Inf_d)$
   
   \If{No. of elements in $Inf_d$ with value equal to $Inf_d[idx] > 1$}
  
  $idx = argmin(Inf_i)$ while $inf_d[idx] = min(inf_d)$
  
  \Return $P_r[idx]$
\caption{ProactiveRoutePlanning($r$,$R'$ $G$ )}
\label{alg:ProactiveRoutePlanning}
\end{algorithm}

\section{Experimental evaluation}\label{sec:exp}

\subsection{Experiment Setup}
\noindent \textbf{Dataset.} We conduct our experiments on a real dataset, consisting of 8 cities in USA (Chicago, Indianapolis, Cincinnati, Dayton, Columbus, Fort Wayne, Cleveland, and Detroit) and 792 CSs within the bounding box of these cities (the area between latitude-longitude $<38.991689, -88.942554>$ to $<42.847614, -81.275585>$). The details of the CSs including their charging rates are from \texttt{afdc.energy.gov}. To the best of our knowledge, there is no existing dataset for inter-city EV trips. Therefore, based on the real trips between these cities~\cite{bts_report}, we generate the origin and destination pair of the EV requests by following the same distribution of origin-destination pairs of real trips happened in January 2021. We denote this request set based on real data as \textbf{RDS}.

To determine different EV properties based on real data, we do as follows. While the average range of EVs is typically 300 km in urban areas, their energy consumption increases by approximately~70\% on highways~\cite{musk_efficiency}. Hence, we consider the driving ranges of EVs as 175 km. For simplicity, we assume uniform driving ranges for all EVs and the energy consumption (i.e., charge lost) is 0.28kW/km. The charging rates of the EVs are randomly chosen from the list in Table~\ref{tab:parameters} to represent diverse charging speeds of EVs in real life. Also, an additional duration of 5 minutes is considered for each charging event to represent real-life delays in the CSs. Departure times follow a Poisson process to reflect different times of the day, utilizing rates from~\cite{xu2019understanding} for low, high, and mid request arrival rates with values of 2, 70, and 40 vehicles per minute, respectively, over durations of 400, 30, and 25 minutes.

\begin{table}[t]
    \centering
    \caption{Parameters}
    \vspace{-0.25cm}
    \begin{tabular}{c|c}
       Parameter & Value \\ \hline
       Number of vehicles & 500, 1K, 1.5K, 2K, 2.5K, 3K, 3.5K, 4K \\
       Length of each trip & 200, 300, 400, 500, 600 (km) \\
       Number of lookaheads & 10, 25, 50, 100 \\
       Timeslot length & 5, 10, 15 \\
    Leave Charging Buckets (LCB) & 50, 75, 100 \\
    Vehicle charging rate & 50, 60, 75, 80, 100 (kWh) \\
    Upper bound limit ($\epsilon$) & 10 (mins) \\
    \end{tabular}
    \label{tab:parameters}
    \vspace{-14pt}
\end{table}

In addition, we generate a second request set with specific properties of the requests to better demonstrate the effect of different parameters. Specifically, this consists of 4000 EV requests with random origin and destinations, all with a given fixed distance between origin and destination within the area mentioned. We later show experiments by varying this distance. The arrival rate of requests is two requests per minute. We denote this request set as \textbf{SDS}.

\vspace{-2pt}
\myparagraph{Evaluation Metrics} We evaluate both the effectiveness and the efficiency under different parameter configurations. For effectiveness we present the `total travel time'. For efficiency, the query time is reported, which is the time required to compute the resulting route after a request is received. 

\vspace{-1pt}
\myparagraph{Baselines} As there is no prior work that directly finds optimal charging path with multiple partial recharging (Section~\ref{sec:rel}), we compare our solution against the closest state-of-the-art CSDB approach~\cite{schoenberg2022reducing}. We also compare the performance of the following different variants of our proposed algorithms: 1) \textbf{\textit{OCP}}: Finds the optimal charging path for each request individually without proactive optimization. 
2) \textbf{\textit{OCP-F}}: Finds the OCP for each request individually without proactive optimization where at each CS, vehicles only get full charge. 3) \textbf{\textit{OCP-PO(n)}}: Finds OCP for each request with proactive optimization that look ahead $n$ future requests.
4) \textbf{\textit{OCP-OC}}: A version of the \textit{OCP} algorithm that selects the path with the minimal charge time. 5) \textbf{\textit{OCP-OCS}}: An alternative version of \textit{OCP} that incorporates a heuristic for calculating the amount of charge a vehicle cannot receive due to having $\Gamma < \eta$ as the unused charging potential. It then selects the path with the lowest unused charging potential.
6) \textbf{\textit{B\&B-NW}}: A branch and bound method that considers zero waiting time at CSs.

\myparagraph{Implementation} All algorithms were implemented in Python 3.9. All experiments were conducted on a Linux server with AMD 1.99GHz CPU and 32GB memory. 

\vspace{-10pt}
\subsection{Performance evaluation}
\vspace{-5pt}

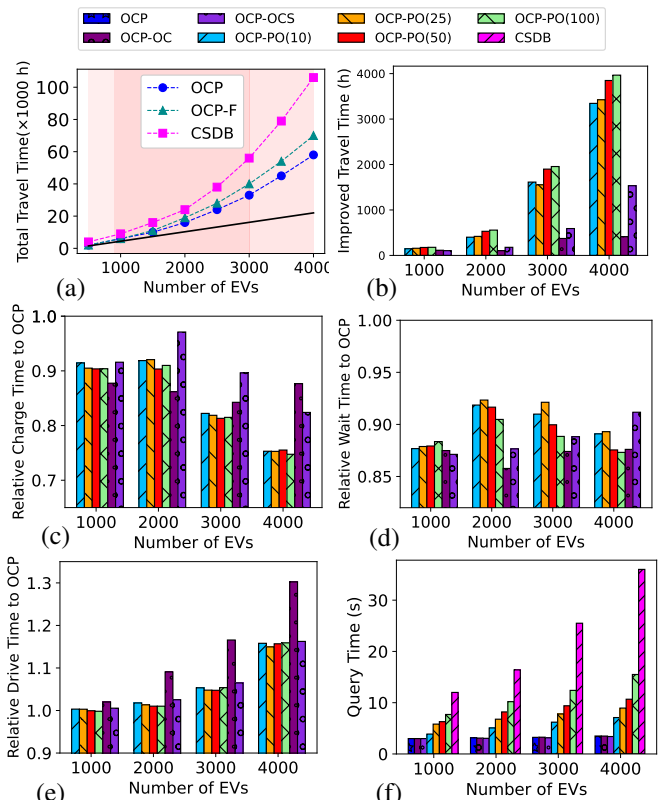
\begin{figure}[h]
  \centering
\hspace{8pt}\subfloat{\includegraphics[width=0.8\linewidth]{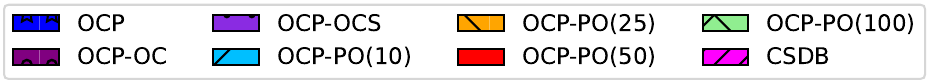}}\vspace{-8pt}
  \\
  \subfloat{
  \begin{tikzpicture}
      \node[inner sep=0] at (0,0) {\includegraphics[width=0.50\linewidth]{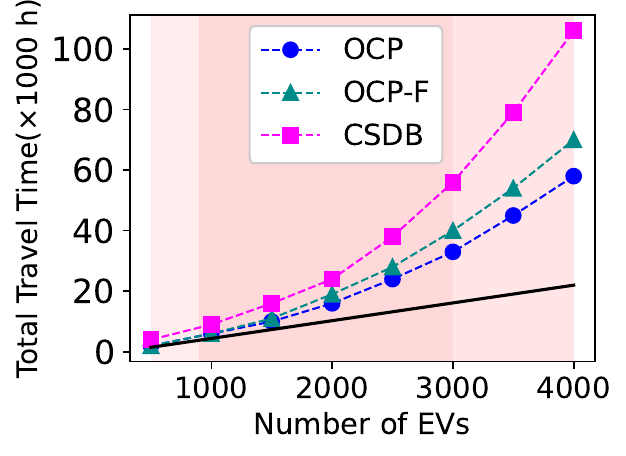}};
      \node[anchor=south west, text=black] at (-1.7,-1.7) {(a)};
    \end{tikzpicture}\label{Fig:TotalTravelTimeRDS_EV}
  }
  \hfill
\subfloat{\hspace{-14pt} \begin{tikzpicture}
      \node[inner sep=0] at (0,0) {\includegraphics[width=0.49\linewidth]{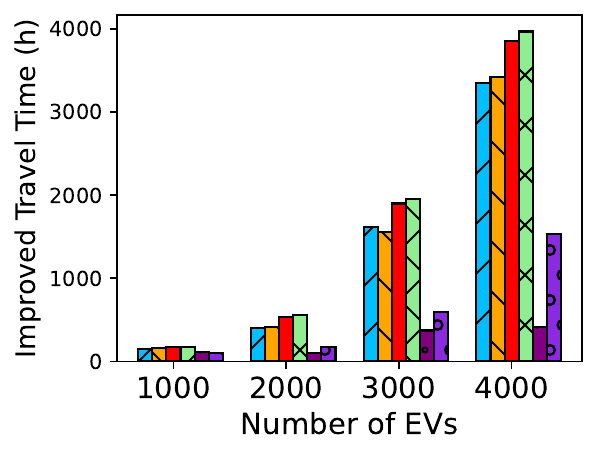}};
      \node[anchor=south west, text=black] at (-1.8,-1.7) {(b)};
    \end{tikzpicture}\label{Fig:ImprovedTravelTime}
  }
  \vspace{-13pt}
 \\
\subfloat{\begin{tikzpicture}
      \node[inner sep=0] at (0,0) {\includegraphics[width=0.49\linewidth]{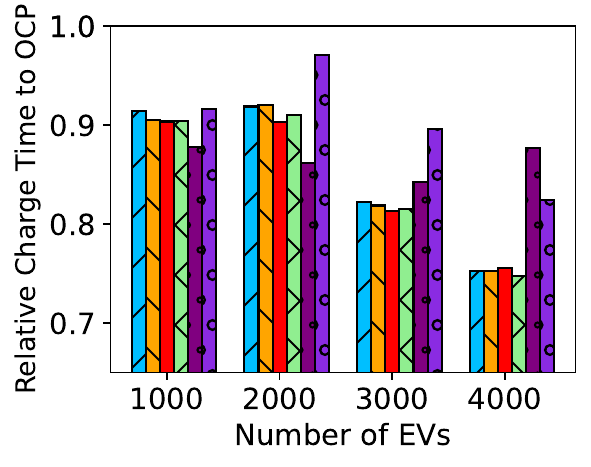}};
      \node[anchor=south west, text=black] at (-1.8,-1.7) {(c)};
    \end{tikzpicture}\label{Fig:RelativeChargeTime}}
\subfloat{\begin{tikzpicture}
      \node[inner sep=0] at (0,0) {\includegraphics[width=0.495\linewidth]{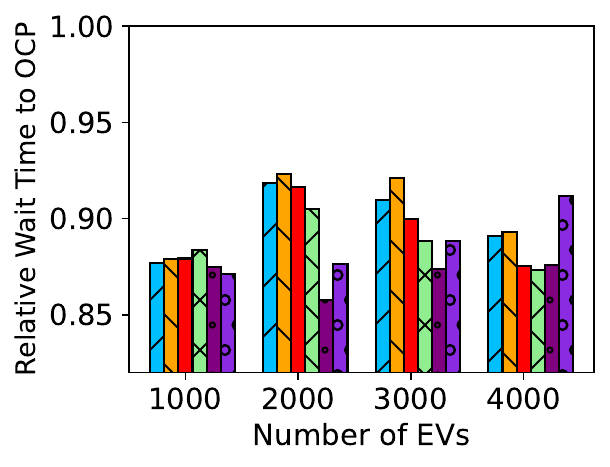}};
      \node[anchor=south west, text=black] at (-1.8,-1.7) {(d)};
    \end{tikzpicture}\label{Fig:RelativeWaitTime}}
\vspace{-13pt}
 \\
\subfloat{\begin{tikzpicture}
      \node[inner sep=0] at (0,0) {\includegraphics[width=0.49\linewidth]{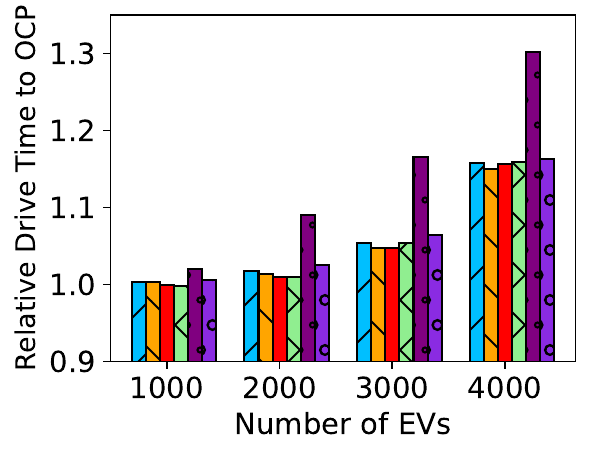}};
      \node[anchor=south west, text=black] at (-1.8,-1.8) {(e)};
    \end{tikzpicture}\label{Fig:RelativeDriveTime}}
\hfill
\subfloat{\begin{tikzpicture}
      \node[inner sep=0] at (0,0) {\includegraphics[width=0.49\linewidth]{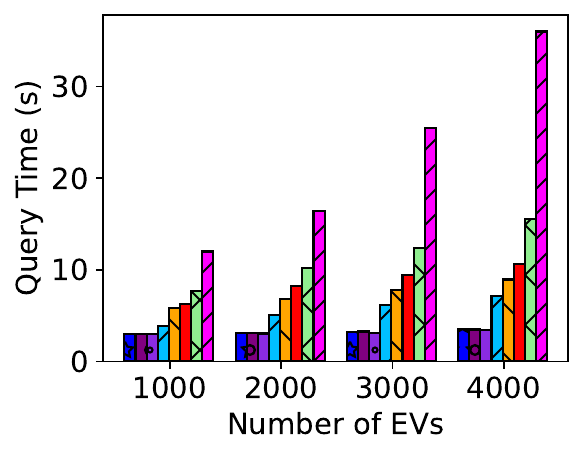}};
      \node[anchor=south west, text=black] at (-1.8,-1.8) {(f)};
    \end{tikzpicture}\label{Fig:QTSDS_EV}}
  \vspace{-13pt}
  \caption{Performance for varying the number of EVs: a) Total travel time comparison with the baseline in RDS b) Improved travel time by proactiveness in RDS c) Relative charge time in RDS d) Relative wait time in RDS e) Relative drive time in RDS f) Mean query time in SDS.  }
  \label{Fig:EV_RDS}
    \vspace{-12pt}
\end{figure}

\begin{figure}
\centering
\subfloat{\includegraphics[width=0.8\linewidth]{figures/query_run_time_legend3.pdf}\label{Fig:TotalTravelTimeRDS}} \\ \vspace{-10pt}
    \subfloat{\begin{tikzpicture}
      \node[inner sep=0] at (0,0) {\includegraphics[width=0.49\linewidth]{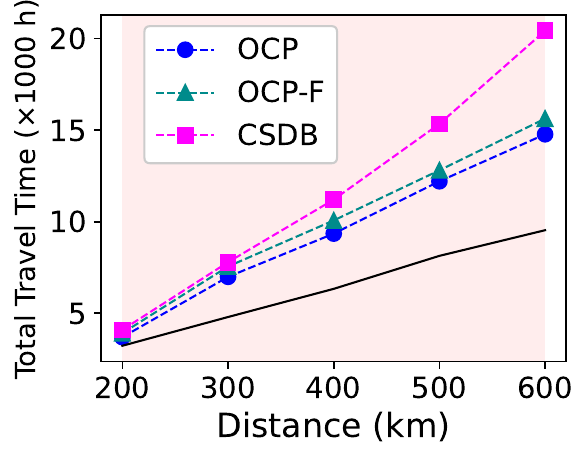}\label{Fig:TotalTravelTimeRDS}};
      \node[anchor=south west, text=black] at (-1.9,-1.8) {(a)};
    \end{tikzpicture}}
  \hfill
  \subfloat{\begin{tikzpicture}
      \node[inner sep=0] at (0,0) {\includegraphics[width=0.49\linewidth]{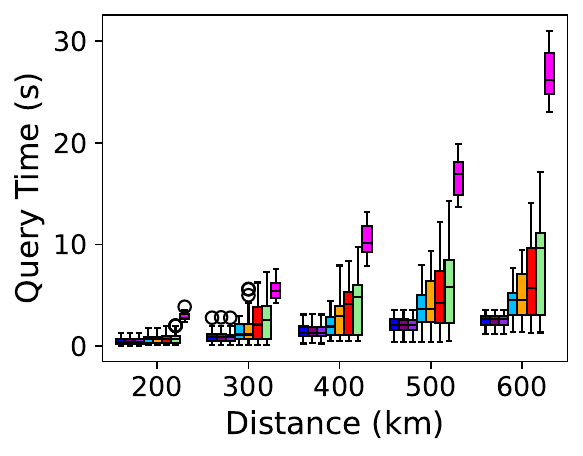}\label{Fig:QueryExecutionTime1000}};
      \node[anchor=south west, text=black] at (-1.8,-1.8) {(b)};
    \end{tikzpicture}}
    \vspace{-8pt}
    \caption{Performance for varying the travel distance in SDS when the number of EVs is 1000: a) Total travel time in SDS b) Mean query time in SDS.}
    \vspace{-10pt}
\end{figure}

\myparagraph{Varying the number of EVs}
As shown in Figure~\ref{Fig:TotalTravelTimeRDS_EV} for RDS, even \textit{OCP-F} that considers full charges only, demonstrates a 33\% superiority compared to \textit{CSDB}. With partial recharging, \textit{OCP} exhibits advantages of 46\% and 19\% over \textit{CSDB} and \textit{OCP-F}, respectively. While these improvements are significant, note the black line depicting \textit{B\&B-NW}, which measures the total time without any wait time. The purpose of this line is to illustrate the \textbf{unavoidable time} (i.e., the time required to drive from origin to destination and the minimum time to charge without running out of battery). If we take the difference between \textit{B\&B-NW} and other algorithms (i.e., excluding the time that cannot be reduced), then OCP outperforms \textit{CSDB} by 57\% for 4000 EV requests, saving the EVs from spending nearly 47,000 hours on the road in total. 

Figure~\ref{Fig:ImprovedTravelTime} shows the improvement in total travel time by proactively considering various values for $n$ future requests as OCP-PO(n). For considering 100 future requests, the total travel time of \textit{OCP-PO(100)} compared to \textit{OCP} is reduced by more than 3900 hours, equivalent to 59 minutes per EV. Also, the results of two baselines \textit{OCP-OC} and \textit{OCP-OCS} are included to demonstrate that the proactive solution is much superior to these simple heuristics. 

We demonstrate the relative charge time, wait time, and drive time of our approaches compared to OCP in Figure~\ref{Fig:RelativeChargeTime}, Figure~\ref{Fig:RelativeWaitTime}, and Figure~\ref{Fig:RelativeDriveTime}. As the proactive solution avoids using the CSs that can be utilized by others, the charge time reduces with the increase of the number of EVs for all OCP-PO(n). While wait time initially rises with the growing number of EVs, it gradually decreases again. The reason is, although more EVs are supposed to increase wait time, choosing the proactive paths gradually reduces queues in CSs for later arrived EVs. The overall improvement in total time comes at a cost of a slight increase in driving time compared to OCP. 

The experiments in SDS exhibit similar trends for varying the number of EVs (not included due to page limit). Since in SDS, all the requests have a given fixed distance from the origin and destination, we present the mean query time in SDS for distance 400 km to better demonstrate the impact of varying the number of EVs on query time (Figure~\ref{Fig:QTSDS_EV}). As shown, CSDB's mean query time exceeds 30 seconds for 4000 requests, escalating with the number of EVs. The mean query time for OCP and the basic heuristics do not change much, and is around 2 seconds. While the benefit of the proactive approaches comes at a cost of increase in the mean query time, they are still 3-4 times less compared to CSDB.

\myparagraph{Varying travel distance} Figure~\ref{Fig:TotalTravelTimeRDS} and \ref{Fig:QueryExecutionTime1000} show the performance in SDS for varying different travel distances. As shown in Figure~\ref{Fig:TotalTravelTimeRDS}, the benefit of OCP is higher for longer trips since more recharges are required, and OCP presents advantages of 20\% and 4\% over CSDB and OCP-F respectively, for queries that extend 600 km. If we exclude the part of the travel time that is unavoidable based on \textit{B\&B-NW}, then OCP outperforms \textit{CSDB} by 50\% for trips of 600 km distance. 

As shown in Figure~\ref{Fig:QueryExecutionTime1000}, the \textit{CSDB} has the longest query time that increases rapidly. The query time of OCP and the simple heuristics do not vary much. The range of the \textit{OCP-PO}'s upper and lower bounds, with varying lookahead numbers of requests, is determined by the fluctuation in the number of influential future requests. A higher number of influential requests leads to longer execution times. 

\begin{figure}[t]
\centering
  \includegraphics[width=0.85\linewidth]{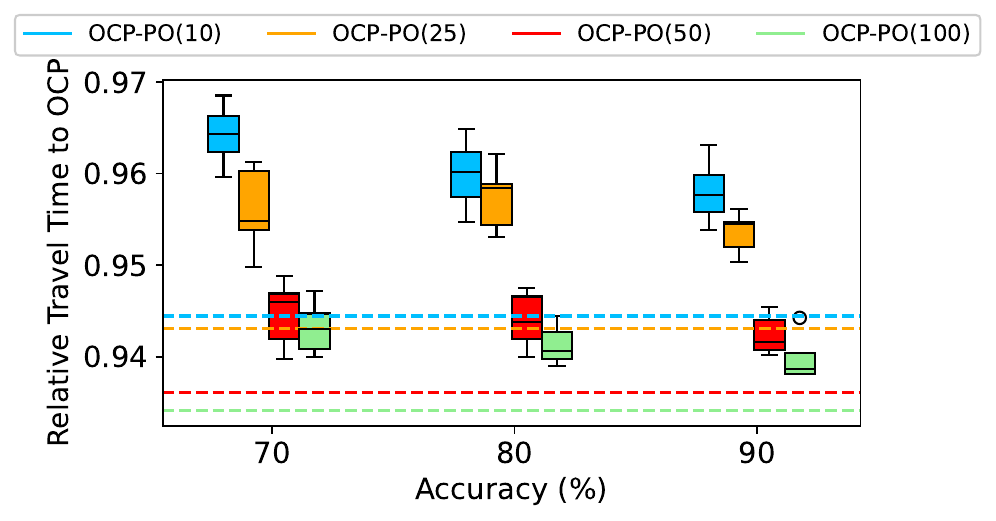}
  \vspace{-10pt}
  \caption{The relative travel time of \textit{OCP-PO(n)} is evaluated by varying the accuracy of the request predictor. The orange, red, blue, and green lines display the results of \textit{OCP-PO(n)} with n 25, 50, 10, and 100, respectively when accuracy is 100\%. }
\label{Fig:AccuracyOfPredictor}
\vspace{-20pt}
\end{figure}

\myparagraph{Impact of accuracy of request predictor} As mentioned earlier, we consider employing a request predictor as a black box. However, due to the inherent unpredictability of certain requests, we assess the OCP's performance under non-deterministic conditions through changing the predictor's accuracy by altering the placement of requests in RDS. For a given accuracy of m\%, we replace 100-m\% of requests. As an example, to illustrate 95\% accuracy, we randomly replace five of the 100 future predicted requests. This adjusted list is used exclusively to provide prediction suggestions, with requests being received by the system as usual. 
Figure~\ref{Fig:AccuracyOfPredictor} presents the results for different \textit{OCP-PO(n)} with varying predictor accuracy for five consecutive runs. The robustness of the algorithm in terms of fluctuating total travel time increases as the number of look ahead requests is increased. The utilization of a predictor with a 90\% accuracy rate can cause a variation of around 0.8\% when employing \textit{OCP-PO(100)}, whereas the fluctuation surpasses 1.7\% for \textit{OCP-PO(10)}.  

\begin{figure}
  \centering
  \subfloat{\includegraphics[width=0.8\linewidth]{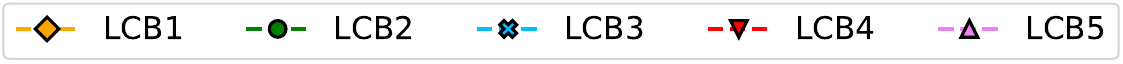}} \\ \vspace{-10pt}
  \subfloat{\begin{tikzpicture}
      \node[inner sep=0] at (0,0) {\includegraphics[width=0.534\linewidth]{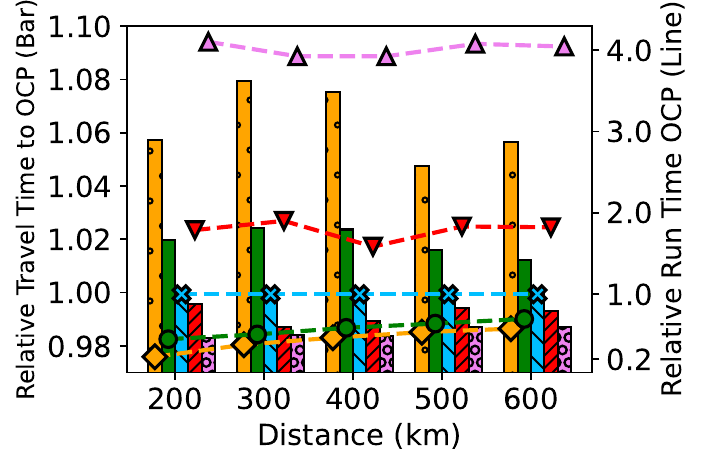}};
      \node[anchor=south west, text=black] at (-1.9,-1.6) {(a)};
    \end{tikzpicture}\label{Fig:RelativeTravelAndQueryTime}}
  \hfill
  \subfloat{\begin{tikzpicture}
      \node[inner sep=0] at (0,0) {\includegraphics[width=0.465\linewidth]{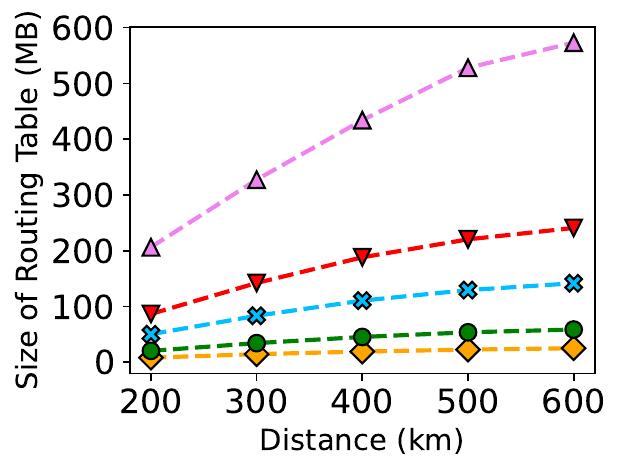}};
      \node[anchor=south west, text=black] at (-1.9,-1.6) {(b)};
    \end{tikzpicture}\label{Fig:SizeOfRoutingTable}}
    \vspace{-12pt}
  \caption{(a) Relative travel time and query time to OCP with LCB3 (b) Size of routing table for 1000 EVs in SDS, where LCB1 = \{100\}, LCB2 = \{50,100\}, LCB3=\{50,75,100\}, LCB4=\{20,40,60,80,100\}, LCB5=\{50,60,70,80,90,100\}.}
  \label{Fig:LCB}
  \vspace{-10pt}
\end{figure}

\myparagraph{Varying LCB} The leaving charging bucket (LCB) represents the SoC values that an EV can recharge to. Figure~\ref{Fig:RelativeTravelAndQueryTime} shows the travel and query time trade-off of different LCBs. Since having more buckets mean more columns in the routing table, we also show the size of the routing table for varying LCB. As more buckets mean more refined granularity of partial recharging, both the relative query time and the size of the routing table are much larger for LCB5=\{50,60,70,80,90,100\} compared to the default LCB3=\{50,75,100\}. However, the relative travel time of LCB5 is only around 98\%. Hence, while having partial recharging significantly improves the performance compared to full recharging, highly fine tuned partial recharging do not bring much additional benefit.

\begin{figure}
\centering
\subfloat{
\includegraphics[width=0.7\linewidth]{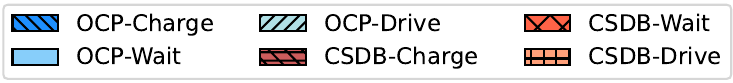}} \\
  \includegraphics[width=0.45\linewidth]{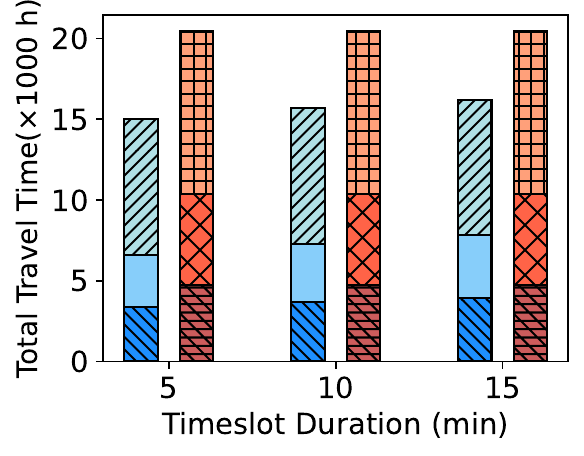}
  \vspace{-13pt}
  \caption{Effect of varying timeslots on total travel time of 1000 EVs in SDS}
  \vspace{-20pt}
\label{Fig:TotalTravelTimeDifferentTimeSlot}
\end{figure}

\myparagraph{Varying timeslot length}
Figure~\ref{Fig:TotalTravelTimeDifferentTimeSlot} displays the travel time for various durations of timeslots in the SDS for 1000 EVs. As the timeslot duration extends from 5 to 15 minutes, the total travel time increases by around 6\%. This increase primarily stems from spending more time at CSs, leading to longer total charge and wait times. However, even with 15-minute timeslots, our proposed algorithm still outperforms the state-of-the-art, as evidenced by the disparity between OCP and CSDB.

\section{Discussion}\label{Discussion}

\vspace{-5pt}
\myparagraph{Generalizability of the model} GTDS provides a general model for the EV charging problem where it is straightforward to incorporate several different variations. Specifically, (1) The travel time between vertices is weights of the edges. Such weights can easily be variables instead of fixed values, to incorporate different travel time at different hours of the day. (2) As shown in literature, the energy consumption (i.e., the resource lost) on the same road can be different for different speed. As the energy consumption is a value associated with each edge, $r(e)$, this can easily be changed based on the speed at that edge. (3) The fluctuation of CSs’ charging rate, influenced by factors like grid status at different times, can be easily captured using time-dependent self-loops. (4) The edges can be directed and have different weights where the travel time from $v_i$ to $v_j$ is different than from $v_j$ to $v_i$. 

\myparagraph{Generalizability of the solution} The proposed solution considers a reservation system at CSs. However, the proposed model and solution is compatible with FCFS queues. Employing the proactive lookahead allows for the effective calculation of waiting time, avoiding potential rerouting in the system, which is the reason for prolonged travel and drive times. 

\section{Conclusion}

In this paper, we propose a novel Proactive EV Route Planning problem and formulate it as an NP-hard graph problem. We present an efficient two-phase solution to find paths with the minimum travel time with multiple partial recharging. We propose influence factor and show that selecting a path with the minimum influence factor can result in minimized travel times overall. Our experiments show that our approach surpasses the current state-of-the-art by 50\% in terms of total travel time. Moreover, our approach effectively decreases query run time, exhibiting improvements of approximately 90\% and 63\% for route planning with and without consideration of influence factor, respectively.   

\vspace{-5pt}
\bibliography{main.bib} 
\bibliographystyle{IEEEtran}
\vspace{-35pt}
\begin{IEEEbiography}[{\includegraphics[width=1in,height=1.25in,clip,keepaspectratio]{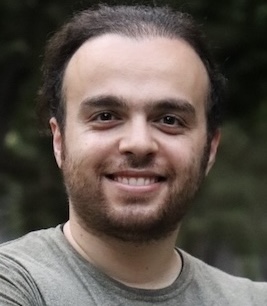}}]{Saeed Nasehi} received his master’s degree in information technology from Sharif University of Technology in 2017 and he is currently a PhD candidate at the University of Melbourne. His research interests include spatial algorithms and location-based services, especially those that intersect with Electric Vehicles. \end{IEEEbiography}
\vspace{-38pt}
\begin{IEEEbiography}[{\includegraphics[width=1in,height=1.2in,clip,keepaspectratio]{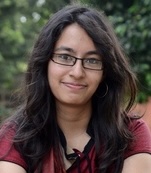}}]{Farhana Choudhury} received her PhD in computer science from RMIT University, Australia in 2018. She is currently a lecturer at the University of Melbourne. Her current research interests include spatial databases, data visualization, trajectory queries, and applying machine learning techniques to solve spatial problems.\end{IEEEbiography}
\vspace{-38pt}
\begin{IEEEbiography}[{\includegraphics[width=1in,height=1.25in,clip,keepaspectratio]{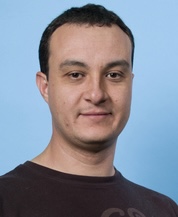}}]{Egemen Tanin} received his PhD from the University of Maryland, College Park. After PhD, he joined The University of Melbourne. His research interests include spatial databases. He has supervised 20 PhD students to completion and has more than 150 articles in his areas of research. He is an Associate Editor of ACM TSAS and he served as the PC Chair for ACM SIGSPATIAL 2011 and 2012. He was elected to serve in many roles for ACM SIGSPATIAL in recent years. He is also the co-founder of ACM SIGSPATIAL Australia and the Founding Editor for ACM SIGSPATIAL Special.\end{IEEEbiography}
\end{document}